\documentclass[lettersize,journal]{IEEEtran}
\usepackage{amsmath,amsfonts}
\usepackage{algorithm}
\usepackage{array}
\usepackage[caption=false,font=normalsize,labelfont=sf,textfont=sf]{subfig}
\usepackage{textcomp}
\usepackage{stfloats}
\usepackage{url}
\usepackage{verbatim}
\usepackage{graphicx}
\usepackage{cite}

\usepackage{hyperref}
\usepackage{booktabs}
\usepackage[edges]{forest}
\usepackage{tikz}
\usetikzlibrary{trees, positioning, shapes.geometric, arrows.meta}
\usepackage{geometry}
\definecolor{LightRed}{rgb}{1,0.92,0.92}
\definecolor{LightOrange}{rgb}{1,0.95,0.88}
\definecolor{LightYellow}{rgb}{1.0,1.0,0.84}
\definecolor{LightGreen}{rgb}{0.9,1.0,0.88}
\definecolor{LightCyan}{rgb}{0.9,1,1}
\definecolor{LightBlue}{rgb}{0.9,0.94,1}
\definecolor{LightIndigo}{rgb}{0.92,0.9,1}
\definecolor{LightMagenta}{rgb}{0.96,0.86,1}
\definecolor{DirtyWhite}{rgb}{0.96,0.96,0.96}
\definecolor{Training}{rgb}{0.9568627450980393, 0.7803921568627451, 0.6901960784313725}
\definecolor{Finetuning}{rgb}{0.6274509803921569, 0.8470588235294118, 0.9372549019607843}
\definecolor{Free}{rgb}{0.596078431372549, 0.8509803921568627, 0.5568627450980392}

\definecolor{LightRed}{rgb}{1,0.92,0.92}
\definecolor{LightOrange}{rgb}{1,0.95,0.88}
\definecolor{LightYellow}{rgb}{1.0,1.0,0.84}
\definecolor{LightGreen}{rgb}{0.9,1.0,0.88}
\definecolor{LightCyan}{rgb}{0.9,1,1}
\definecolor{LightBlue}{rgb}{0.9,0.94,1}
\definecolor{LightIndigo}{rgb}{0.92,0.9,1}
\definecolor{LightMagenta}{rgb}{0.96,0.86,1}
\definecolor{DirtyWhite}{rgb}{0.96,0.96,0.96}
\usepackage{multirow}
\usepackage{colortbl}

\usepackage{xcolor}
\hypersetup{
    colorlinks=true,
    linkcolor=red,
    urlcolor=magenta,
    citecolor=cyan
}
\usepackage{ textcomp }
\usepackage[utf8]{inputenc}
\usepackage{enumitem}
\usepackage{nicefrac}
\usepackage{xurl}
\usepackage{tabularray}
\usepackage{epstopdf}
\usepackage{makecell}
\usepackage{threeparttable}
\usepackage{longtable}
\usepackage{multirow}
\usepackage{tabularx}
\usepackage{tabulary}
\usepackage{tablefootnote}
\usepackage{array}
\usepackage{dsfont}
\usepackage{epstopdf}
\usepackage{graphicx}
\usepackage{caption}
\usepackage{color}
\usepackage{mathtools}
\usepackage{url}
\usepackage{bm}
\usepackage{graphicx}
\usepackage{enumitem}
\usepackage{xcolor}
\usepackage{algpseudocode}
\usepackage{titlesec}

\usepackage{amssymb} 
\begin{document}

\title{From Fragment to One Piece: A Survey on AI-Driven Graphic Design}

\author{Xingxing Zou, Wen Zhang, Nanxuan Zhao
\thanks{Xingxing Zou is with the Hong Kong Polytechnic University, Hong Kong SAR, China. Wen Zhang is with Snap Inc, CALIFORNIA, United States Nanxuan Zhao is with Adobe Inc, CALIFORNIA, United States. (Corresponding author: Xingxing Zou, email: xingxing.zou@polyu.edu.hk).} 
}

\markboth{Journal of \LaTeX\ Class Files,~Vol.~14, No.~8, August~2021}%
{Shell \MakeLowercase{\textit{et al.}}: A Sample Article Using IEEEtran.cls for IEEE Journals}


\maketitle

\begin{abstract}
This survey provides a comprehensive overview of the advancements in Artificial Intelligence in Graphic Design (AIGD), focusing on integrating AI techniques to support design interpretation and enhance the creative process. We categorize the field into two primary directions: perception tasks, which involve understanding and analyzing design elements, and generation tasks, which focus on creating new design elements and layouts. The survey covers various subtasks, including visual element perception and generation, aesthetic and semantic understanding, layout analysis, and generation. We highlight the role of large language models and multimodal approaches in bridging the gap between localized visual features and global design intent. Despite significant progress, challenges remain to understanding human intent, ensuring interpretability, and maintaining control over multilayered compositions. This survey serves as a guide for researchers, providing information on the current state of AIGD and potential future directions\footnote{https://github.com/zhangtianer521/excellent\_Intelligent\_graphic\_design}.
\end{abstract}


\section{Introduction}
\IEEEPARstart{F}{orecasts} by McKinsey and PricewaterhouseCoopers suggest that generative AI in graphic design could potentially contribute more than \$8 trillion to the global economy by 2030. Research interest in this area has continuously increased.

\begin{figure*}[h!]
  \begin{center}
  \includegraphics[width=1\linewidth]{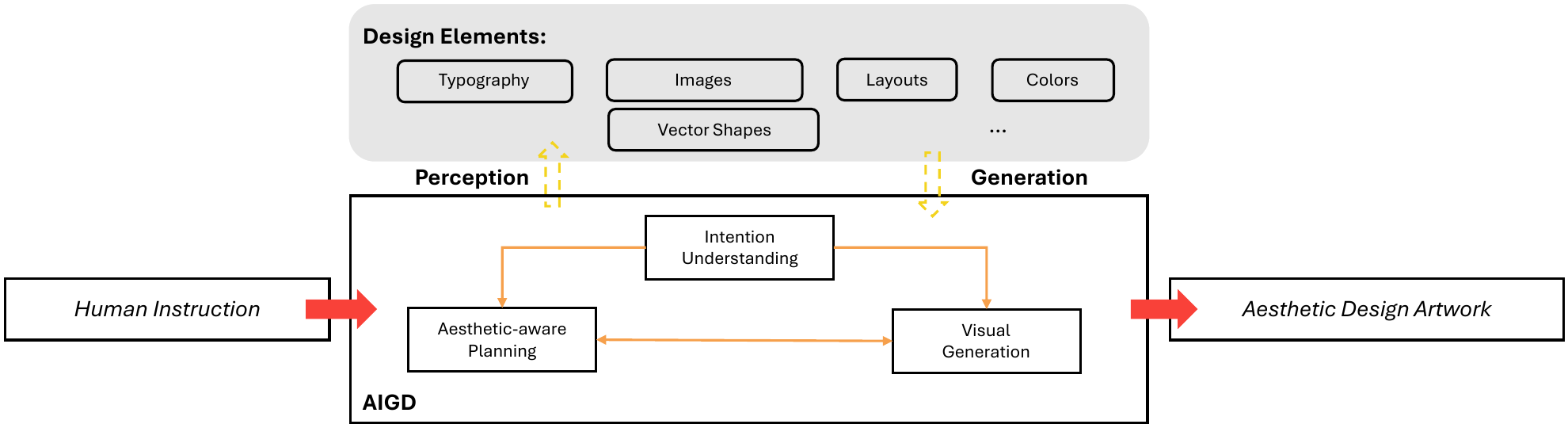}
  \end{center}
  \vspace{-3.5mm}
  \caption{General pipeline of Artificial Intelligence in Graphic Design (AIGD).} 
  \label{fig:aigd-overview}
  \vspace{-5.5mm}
\end{figure*}

The academic evolution of Artificial Intelligence in Graphic Design (AIGD) reveals two distinct phases. Early efforts focused on decomposing design tasks into atomic components—typography generation \cite{jiang2019visual}, layout optimization \cite{weng2023learn,feng2024layoutgpt}, and color palette recommendation \cite{tan2016decomposing,tan2018efficient,wang2019improved,zhang2016colorful,vitoria2020chromagan}—employing specialized models for each subtask. While effective for generating individual elements, this decompositional approach introduced systemic fragmentation that persists in current research\cite{huang2023survey}. Recent breakthroughs in large-scale text models have catalyzed the evolution of generative visual models~\cite{liu2024visual, hu2024instruct}. Within graphic design~\cite{epstein2023art}, as illustrated in Fig.~\ref{fig:aigd-overview}, this progress reflects a paradigm shift: from optimizing isolated elements (e.g., typography, images, vector shapes, layouts, and colors) to holistic creative systems capable of maintaining aesthetic consistency across entire design workflows—from human instruction to final artwork. Tools like Adobe Firefly\footnote{\url{https://www.adobe.com/}} also exemplify this trend, highlighting the transformative potential.

Recent surveys in graphic design have explored various dimensions of the field. \cite{tian2022survey} analyze vector graphics through mathematical foundations and content creation stages; \cite{shi2023intelligent} delve into layout generation aesthetics and technologies; \cite{huang2023survey} provide a taxonomy of graphic design intelligence. \cite{liu2024intelligent} review graphic layout generation focusing on implementation and interactivity, while ~\cite{tang2024s} investigate challenges and future functional needs for AI-generated image tools in graphic design through designer interviews.

\textbf{Scope of the Reviews.} The multimodal era has witnessed attempts at cross-modal integration through vision-language models \cite{zhang2024vision,li2022blip}. However, these have largely failed to bridge the semantic gap between localized visual features and global design intent \cite{lin2024designprobe}. Recent advances in LLM-driven design systems \cite{lin2024designprobe,cheng2024graphicllm} suggest a convergence path, where generative processes are guided by explicit design rationale encoded in latent spaces \cite{xiao2024omnigen,zhou2024transfusion}. With this understanding, we approach AIGD from a fresh and intuitive angle, \textbf{centring our discussion on design understanding and creativity.} Unlike earlier studies that often zero in on isolated technical improvements, we take a holistic view of the entire design process—bridging the gap between concept and creation, intending to help readers quickly grasp the scope of AIGD and see how the latest AI advancements can enhance the creativity of graphic design. 

\begin{figure*}[t]
    \begin{center}
    \includegraphics[width=1\linewidth]{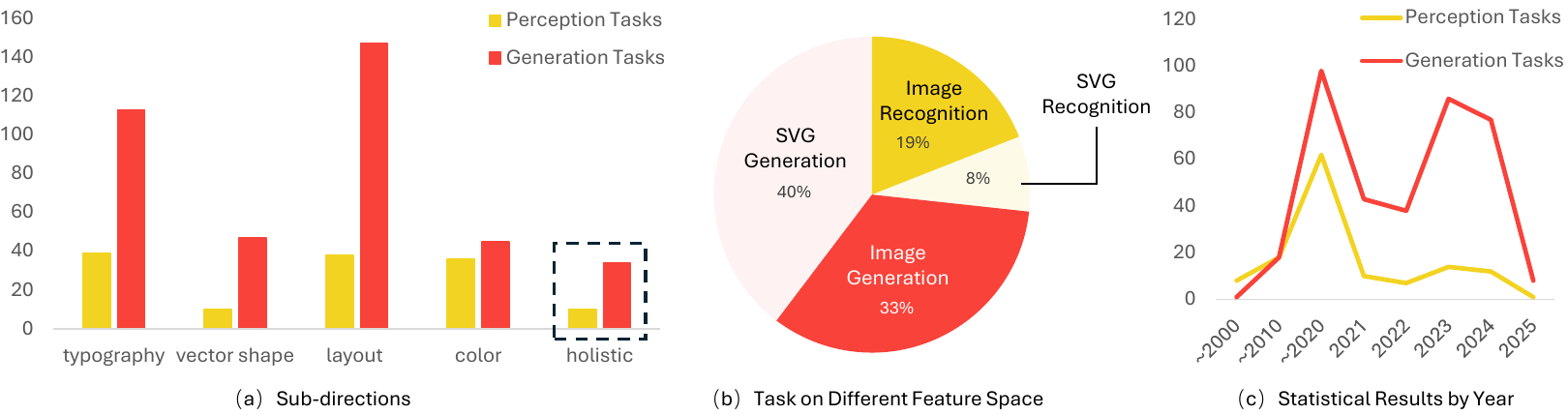}
    \end{center}
    \vspace{-4.5mm}
    \caption{Overview of research publications in AIGD. (a) sub-directions, (b) task on different feature space, and (c) statistical results by years. Holistic indicates systems capable to process multiple or all design elements.} 
    \label{fig:taxonomy}
    \vspace{-5.5mm}
  \end{figure*}

To this end, we conducted an extensive survey and in-depth analysis focusing on AIGD. By categorizing AIGD research through the dual lenses of design semantics (e.g., visual hierarchy, typography, color theory) and creative workflows (e.g., ideation, iteration, refinement), we provide a unified framework to evaluate progress in this domain. Unlike prior fragmented analyses, our survey explicitly examines how AI models interpret and generate meaningful design artifacts—whether raster or vector graphics—while preserving artistic intent and adaptability. Our survey reviewed approximately 500 research papers and categorized them into two main directions: cognitive and generative.  Within these categories, we explored subtasks and methods associated with four key design elements: non-text objects, text characters, aesthetic elements, and layout. Fig.~\ref{fig:taxonomy} intuitively represents the statistical distribution between cognitive and generative research. Our statistical results reveal that:
(1) Current research predominantly focuses on processing individual subtasks, with comprehensive studies on graphic design as a holistic task emerging only since 2023.
(2) Although raster images are more prevalent in general research, vector images attract significant attention in AIGD due to their characteristics, such as lossless scaling.
(3) Font generation and layout generation for raster images are more common, whereas bitmap images are more frequently studied for generating non-text objects. This discrepancy may be attributed to the fact that most research on raster image generation does not explicitly target graphic design, leading to domain incompatibility.
(4) There is significant enthusiasm among researchers for this field, with 2010 marking a critical juncture between recognition and generation tasks. Since then, interest in generation tasks has consistently outpaced cognitive tasks, notably increasing in 2022.

  Graphic design is inherently comprehensive. In this survey, we endeavour to comprehensively review the methods involved in various subtasks and discuss the challenges, unresolved issues, and future research directions. The remainder of this article is organized as follows: We first present the background of graphic design in Sec.\ref{pro}, introducing relevant basic concepts, involved subtasks, and the organizational structure of this paper, divided into cognitive and generation tasks. Sec.\ref{sec:cognition tasks} discusses cognitive tasks while Sec.\ref{sec:generation tasks} focuses on generation tasks. We limit our scope to graphic design or closely related domains when considering research papers. Due to the breadth of content involved, we provide highly condensed information to facilitate reader comprehension and identify research branches. Finally, from a holistic graphic design perspective rather than individual subtasks, we examine the state-of-the-art progress in the Multimodal Large Language Model era, existing challenges, and potential future trends in Sec.\ref{cha}.

\section{Background}
\label{pro}
Graphic design aims to deliver information clearly while presenting it in an appealing visual way~\cite{meggs1992type}. It involves design elements, including non-text objects (images and vector shapes) and text characters (typography), to create aesthetic narratives\cite{meggs1992type} through visual harmony layout and aesthetic elements, particularly colors. Graphic design typically employs two principal data types: raster and vector images.

\begin{itemize}
    \item \textbf{Raster Image.} is a two-dimensional array storing pixel values, with the pixel as its fundamental unit influenced by resolution.
    
    \item \textbf{Scalable Vector Graphics (SVG).} uses mathematical descriptions to record content, such as parameters to draw straight lines.
    \end{itemize}
    
    We first define the problem of AIGD under a unified mathematical formulation. Let \( E = T \cup O \) represent the set of all design elements, where \( T \) is the set of text elements (typography) and \( O \) is the set of non-text elements (images or vector shapes). Each element \( e \in E \) in the design is described by two kinds of features: 1) its attribute vector \( a_e \), which may include style, size, color, etc. 2) its design content \(c_e\), such as visual context of objects. Therefore, a candidate design artifact is a set of element features, \(D(A=\cup{a_e},C=\cup{c_e})\). The objective of the design is to find an optimal \(D(A,C)\) which maximize the following function:
    \begin{equation}
    D(A, C) = \arg \max_{A, C} V(L(A), C \mid I )
    \label{equ1}
    \end{equation}
    where \( V(.) \) measures the aesthetic value of design \(D\) under the constraint of user intention \(I\). \(L(A)\) is the layout of design elements, 
    \begin{equation}
    L(A) = h(\tilde{A}|C)\ge \tau
    \label{equ2}
    \end{equation}
    which follows the certain design principles (\(h(.)\)) providing best harmonic settings among design objects. \(\tau\) is the minimal harmonic score. In Eq.\ref{equ2}, not all attributes are applicable for layout. Typically, a subset \(\tilde{A}=(p_i, s_i, \theta_i)\) is the core parameters, where $p_i \in \mathbb{R}^2$ represents the \textbf{position} coordinates of element $e_i$ (for a two-dimensional layout), $s_i \in \mathbb{R}^+$ represents the \textbf{size} of an element $e_i$ (e.g., width and height), and $\theta_i \in S$ represents other states of element $e_i$ (e.g., rotation angle, transpose, etc.). Consider the human centric nature of graphic design, human intention \(I\) steers the exploration of \(D(A,C)\). However, interpretation of human intention is complex and requires consideration of individual preference. It is provided as a part of system inputs, in terms of multi-modal instructions, reference images or templates.
    
    Ideally, it is anticipated that Eq.~\ref{equ1} could be resolved within a unified pipeline. The major components would be capable of comprehending design intentions \(I\), acquiring basic relevant elements \( E_{basic}\), orchestrating graphic layouts \( L(A) \), and ensuring the visual harmony of the produced outcomes \( V(D) \). Recent studies published between 2023 and 2024 have explored the potential of LLMs in graphic design, as evidenced by works such as those by Dou et al. (2024)~\cite{dou2024hierarchical}, Huang et al. (2024)~\cite{huang2024graphimind}, and others~\cite{lin2024designprob,ding2023designgpt,weng2024desigen}. However, these efforts are still in the early stages and have yet to achieve a deep understanding and professional generation.

    \begin{figure*}[t]
        \begin{center}
        \includegraphics[width=1\linewidth]{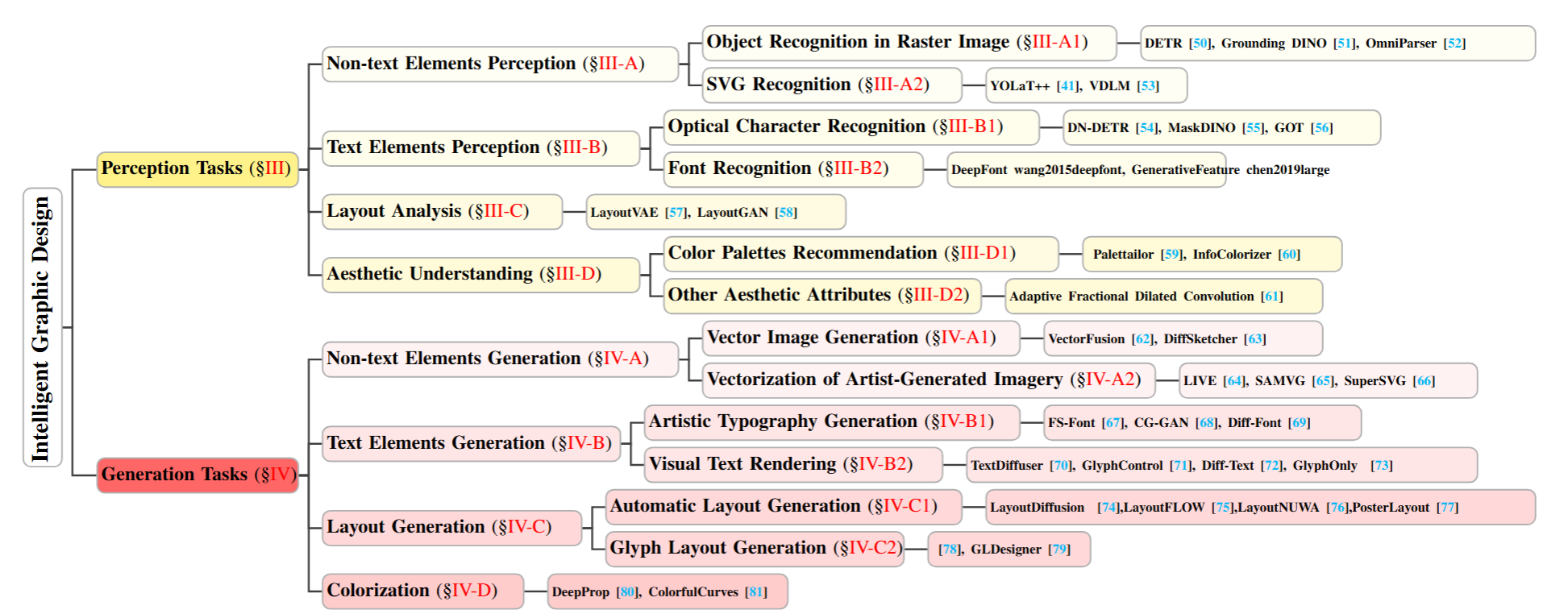}
        \end{center}
        \vspace{-2.5mm}
        \caption{Literature taxonomy of AIGD.} 
        \label{fig:1}
        \vspace{-4.5mm}
      \end{figure*}

    The primary goal of graphic design is to deliver information. While visual elements, particularly images and vector shapes, enhance aesthetic appeal, text elements directly communicate the design's theme and play a more central role in information delivery. This relative importance is reflected in statistical data from the field, indicating that 31\% of tasks are text-related, while only 12\% focus on non-text elements, including tasks that involve both types of content. Text in graphic design encompasses various attributes such as fonts, glyphs, artistic styles, and semantic typography, all crucial for effective communication. Another significant factor is the meticulous arrangement of these elements. Key messages in the text should be highlighted, and design considerations like adequate white space, optimal contrast, and visual balance are essential for a well-managed layout. Research specifically targeting these considerations accounts for more than 38\% of studies in this area. Additionally, achieving aesthetic appeal through techniques like color harmony is also critical, representing 13\% of research interests. Due to the complex nature of these tasks, much of the existing research has been divided into distinct sub-directions, primarily categorized into two areas: \textbf{Perception Tasks}\cite{luo2020learning,o1993document,long2022towards,cheng2023m6doc,luo2024layoutllm,chen2024rodla,zhang2024m2doc,chen2024fusion} and \textbf{Generation Tasks}\cite{kong2022look,he2024diff,tang2022few,chen2020automatic,wang2023cf}.

\section{Perception Tasks}
\label{sec:cognition tasks}
Understanding design intent is the first step towards AIGD, which requires a model with a basic knowledge of graphic design principles. Therefore, this section presents the methodology evolution separately for each sub-task, including non-text element perception in Sec.~\ref{sec:non text re}, text element perception in Sec.~\ref{sec:text re}, layout analysis in Sec.~\ref{sec:layout re}, and aesthetic understanding in Sec.~\ref{sec:aes re}. As two main types of data are primarily used in graphic design.

\vspace{-3mm}
\subsection{Non-text Element Perception}
\label{sec:non text re}
\vspace{-1mm}
\subsubsection{Object Recognition in Raster Image}
\label{sec:non text re ras}
Numerous comprehensive surveys and reviews have documented advances in non-text object recognition~\cite{cheng2023towards}. Building upon the areas not extensively covered by these surveys, recent progress in Multimodal Large Language Models (MLLMs)\cite{chen2022visualgpt} has significantly expanded the capabilities of LLMs to process and interpret text and visual data. These MLLMs have demonstrated remarkable proficiency in vision-language tasks, such as image captioning and visual question answering. Furthermore, contemporary research has begun to explore the potential of using textual output from LLM to steer external vision expert models to perform a variety of vision-centric tasks~\cite{wan2024omniparser}. In object detection, such expert models include systems such as DETR~\cite{carion2020end}, which are designed to improve the accuracy and efficiency of detecting and interpreting visual objects.

\subsubsection{SVG Recognition}
\label{sec:non text re vec}
Traditional methods employ rule-based graph-matching techniques, such as visibility graphs \cite{locteau2007symbol} and attributed relational graphs \cite{ramel2000structural}. YOLaT\cite{jiang2021recognizing} first proposed a learning-based method that represents vector graphics using graphs based on the Bézier Curve, where object detection is conducted based on the predictions of a Graph Neural Network (GNN). However, this work models only in a flat GNN architecture with vertices as nodes, ignoring the higher-level information of vector data. The follow-up work by \cite{dou2024hierarchical}, YOLaT++, learns multi-level abstraction features from primitive shapes to curves and points. They also provide a new dataset for chart-based vector graphics detection and chart understanding, which includes vector graphics, raster graphics, annotations, and raw data for creating these vector charts.

\vspace{-3mm}
\subsection{Text Element Perception}
\vspace{-1mm}
\label{sec:text re}
\subsubsection{Optical Character Recognition (OCR)} 
\label{sec:text re ra}

To facilitate text recognition, it is essential to locate the text area within an image~\cite{bi2024text}. Popular text detection algorithms can be broadly categorized into regression-based, segmentation-based, and Detection Transformers. Regression-based algorithms draw from general object detection methods, which treat text detection as a unique scenario within target detection, such as TextBoxes~\cite{liao2017textboxes} (based on the Single Shot Multi-box Detector (SSD)~\cite{liu2016ssd}), CTPN~\cite{tian2016detecting} (based on Faster R-CNN~\cite{ren2016faster}), among others. 
ABCNet~\cite{liu2021abcnet} is the first to introduce Bezier curve control points for arbitrary-shaped texts. On the other hand, segmentation-based algorithms, inspired by Mask R-CNN~\cite{he2017mask}, have significantly improved text detection across various scenes and shapes but entail complex post-processing, speed issues, and challenges in detecting overlapping text. DETR~\cite{carion2020end} represents a pioneering model introducing a fully end-to-end transformer-based paradigm.  However, DETR’s training convergence and feature resolution limitations have hindered its competitiveness compared to traditional detectors. Other variants include Conditional-DETR~\cite{meng2021conditional}, Anchor-DETR~\cite{wang2022anchor}. Furthermore, approaches like DN-DETR~\cite{li2022dn} and MaskDINO~\cite{li2023mask} concentrate on label assignment strategies, significantly improving matching stability.

Once text is detected, text recognition algorithms identify the content within the detected areas. It is typically divided into regular and irregular text recognition based on the shape of the text. Regular text includes printed fonts and scanned text, whereas irregular text often appears non-horizontal and may exhibit bending, occlusion, and blurring. Historically, the mainstream approach involved segmentation and single-unit recognition, utilizing connected domain analysis to identify potential text segmentation points. Post-2016, the focus shifted to text line recognition. Early works using DNNs as feature extractors for scene text recognition include \cite{jaderberg2016reading}. However, many texts in natural scenes have arbitrary shapes and layouts, making it difficult to transform them into horizontal texts through the proposed interpolation methods. To this end, the later studies focus on identifying granular-level elements, such as characters, and semantically encoding their relationships to enhance the recognition of irregular text~\cite{liu2018char,cheng2018aon,li2019show}. Many recent works have introduced a growing trend of generative models into scene text recognition. \cite{liu2018synthetically} proposed transforming the entire scene text image into corresponding horizontally written canonical glyphs to promote feature learning. Through their experiments, the guidance of canonical glyph forms proved effective for feature learning in STR.

\subsubsection{Font Recognition}
\label{sec:text re ve}
\begin{figure*}[t]
  \begin{center}
  \includegraphics[width=1\linewidth]{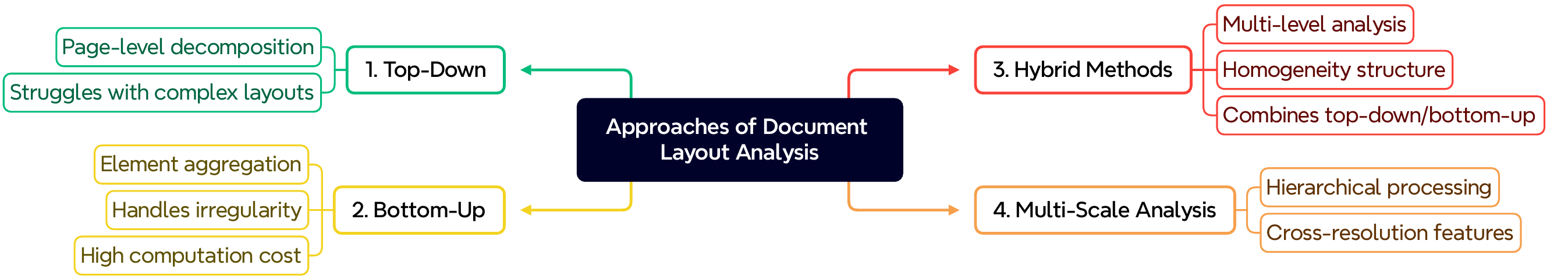}
  \end{center}
  \vspace{-3.5mm}
  \caption{Overview of methods for layout analysis task.} 
  \label{figLAF}
  \vspace{-5.5mm}
\end{figure*}
Fonts are typically designed with unique characteristics, such as stroke width, serifs, aspect ratio, spacing or slant/Italicization~\cite{chen2019large}. Early font recognition works attempted to recognize fonts via these artificial font features~\cite{zhu2001font,chen2014large}. While artificial font features worked reasonably well in controlled scenarios, they faced several challenges: 1. font variability: Fonts with subtle differences in design could be difficult to differentiate using simple features. 2. noise in input data: Scanned documents or degraded images introduced noise that could distort features like stroke width or spacing. 3. handwritten vs. printed fonts: Artificial features were less effective for recognizing handwritten or highly stylized fonts. 4. limited scalability: Adding new fonts required manually defining additional rules or features. As the learning-based method became popular, Wang et al. built a Convolutional Neural Network with domain adaptation techniques for font recognition, applying deep neural networks to font recognition for the first time~\cite{wang2015deepfont}. This method was followed by Bharath et al., who utilized SVM for English font recognition, focusing on character image distances~\cite{bharath2017font}. The research was further expanded by Liu et al., who introduced a multi-task adversarial network for Japanese fonts, employing a GAN to preprocess scene text images prior to recognition~\cite{liu2018multi}. The introduction of the FontCLIP latent space further expands the possibilities for font selection using out-of-domain attributes and scripts, improving flexibility~\cite{sun2024extending}.

\vspace{-3mm}
\subsection{Layout Analysis}
\vspace{-1mm}
\label{sec:layout re}
Layout is composed of visual elements, typically characterized by properties such as type and position. As shown in Fig.~\ref{figLAF}, traditional approaches use hand-crafted features to represent layout. For instance, Stoffel et al. designed features related to position, spacing, and font styles for document structure analysis~\cite{stoffel2010enhancing}. Some methods employ neural networks, such as Transformer~\cite{vaswani2017attention} and Faster-RCNN~\cite{ren2016faster}, to encode layouts into low-dimensional continuous representations, showing promising results~\cite{patil2020read}.

The layout analysis of graphic design shares the common foundation of other aesthetic-aware layout analyses, e.g., Document Layout Analysis (DLA), where domain knowledge can be easily transferred to all general graphic design. DLA methodologies include bottom-up, top-down, hybrid, and multi-scale approaches. The top-down approach starts with each page as a single large block, which is then subdivided into smaller sections, but struggles with complex layouts~\cite{sun2005page}. The bottom-up approach, starting at the granular level and aggregating adjacent elements into larger blocks, handles irregular layouts well but can be computationally demanding~\cite{agrawal2009voronoi++,simon1997fast}. Hybrid methods combine these approaches and utilize a multi-level, homogeneity structure to improve layout analysis~\cite{tran2016page}. The adoption of CNNs has shifted DLA towards models that extract features directly from document pixels, addressing shortcomings of traditional methods. Early CNN models focused on textural features for segmenting segments but were less effective with elements like tables~\cite{vil2013algorithm}. Recent developments like Gruning's ARU-Net and Xu et al.'s multi-task FCN improve text line segmentation and contour detection~\cite{gruning2019two,xu2018multi}. These innovations emphasize the integration of semantic interpretation in DLA, highlighting the importance of understanding semantic relationships between document components. \cite{luo2024local} highlights the importance of contextual relevance in element placement.

\vspace{-3mm}
\subsection{Aesthetic Understanding}
\vspace{-1mm}
\label{sec:aes re}

The field of aesthetic understanding in graphic design has evolved from manual feature engineering to AI-driven multimodal systems~\cite{kong2022aesthetics++}. Early approaches relied on handcrafted color metrics and rule-based harmony models, while modern methods leverage deep learning (GANs, VAEs, Transformers) for context-aware palette generation and personalized recommendations. Concurrently, aesthetic assessment has transitioned from spatial/statistical analysis to neural architectures that model emotional impact and user preferences. This paradigm shift enables unified systems addressing both functional requirements (color discrimination) and affective dimensions (emotional resonance) across infographics, marketing materials, and interactive interfaces.

\subsubsection{Color Palettes Recommendation}
\label{sec:aes re color}
Initial color recommendation systems relied heavily on manual feature extraction and regression models. For instance, colour Sommelier \cite{son2015color} introduced a harmony rating algorithm based on community-generated palettes, allowing users to iteratively select harmonious colour schemes.  However, these methods often overlooked the semantic meanings of colors and incorporated less critical features, leading to suboptimal predictions. The advent of deep learning marked a significant paradigm shift in color recommendation systems. Neural networks began to learn color feature representations from image color histograms, classifying images according to predefined categories. Early efforts included the use of neural networks on predefined color palettes tailored for specific themes, such as magazine cover design~\cite{jahanian2013recommendation, yang2016automatic,maheshwari2021generating, bahng2018coloring}.

Recent studies have focused on recommending color palettes for information visualizations and statistical graphics, such as scatterplots and bar charts~\cite{lu2020palettailor,yuan2021infocolorizer}. Beyond simple infographics, researchers have explored color palette recommendations for more complex visual designs, such as advertising posters and magazine covers. Yuan et al. \cite{yuan2021infocolorizer} implemented a Variational AutoEncoder with Arbitrary Conditioning (VAEAC) to dynamically suggest colors for various infographic elements. The latest research in color recommendation increasingly focuses on generative models and region-specific recommendations. \cite{qiu2022intelligent, qiu2023color}developed a Transformer-based masked color model for specific regions on landing pages and vector graphic documents. \cite{kikuchi2023generative} utilized maximum likelihood estimation and conditional variational autoencoders within a Transformer framework to recommend text and background colors for e-commerce mobile web pages.

\subsubsection{Other Aesthetic Attributes}
\label{sec:aes re other}
Aesthetic visual quality assessment advances from traditional handcrafted feature extraction methods to sophisticated deep-learning approaches, with the trend from standardised assessment to more diverse and personalized evaluation. Early aesthetic visual quality assessment methods focused on extracting handcrafted features from images~\cite{ke2006design,wong2009saliency}. The introduction of models that incorporated human-describable attributes marked a significant advancement. \cite{dhar2011high} introduced a model that connected technical analysis with human perceptions, including elements such as composition, illumination, and content. Obrador et al.\cite{obrador2012towards} evaluated photographs based on features like simplicity and visual balance. \cite{reinecke2013predicting} developed models to predict users' first aesthetic impressions of websites, based on visual complexity and colorfulness. With the advent of deep learning, the field underwent a transformative change. Lu et al.\cite{lu2014rapid, lu2015deep} utilized dual-column CNNs and a Deep Multi-Patch Aggregation Network (DMA-Net) to encode global image layouts better, significantly advancing the classification and understanding of aesthetic qualities. Recent studies have focused on personalized image aesthetics, exploring how users' social behaviors and personal perceptions influence their aesthetic judgments. Cui et al.\cite{cui2020social,cui2020personalized} addressed user-centric aesthetic assessment analysis. Chen et al.\cite{chen2020adaptive} introduced the Adaptive Fractional Dilated Convolution to maintain the original aspect ratios and composition of images.

\vspace{-3mm}
\subsection{Summary}
\vspace{-1mm}
\label{sum1}
The evolution of the visual cognition framework is illustrated by the transition from traditional methods to the adoption of deep learning techniques such as CNNs, subsequently incorporating GANs, Transformers, and, currently, LLMs. Similarly, although each subtask within AIGD progresses independently,  their overall development trajectories align consistently with this. Another notable trend is research on vector images remains relatively sparse compared to raster images. Most studies on vector cognition have focused on SVG recognition, representing earlier efforts in the field. However, recent statistics indicate a growing interest in vector image research. This surge is attributed to the nature of vector representations, which are highly conducive to integration with LLMs for enhanced understanding and reasoning.

In text recognition, the main challenges addressed are OCR and font recognition. General approaches include techniques such as Faster R-CNN, text line or single character segmentation, and the input ranges from handwritten notes to natural scene images. The recognition process faces several challenges, including text distortion due to perspective changes, small text scale, stylized fonts, various font sizes and styles, decorative elements, multilingual text, image blur, and poor lighting conditions. Font recognition is another crucial aspect of text-related cognition and plays an essential role in graphic design where font styles are vital. However, the diversity of font styles poses a significant challenge to creating a comprehensive dataset, including unique styles such as italics and bold. This makes it difficult for models to learn to recognize diverse fonts. 

In addition to text elements, layout analysis, especially in document structure, has received increasing attention from researchers. This analysis is a precursor to OCR, classifying and recognizing different elements in a document, such as text, images, tables, and titles. Recent research has made significant progress through large-scale language model-based tools such as LayoutLM, UDOP, and LiLT, which leverage multimodal Transformer encoders pre-trained and fine-tuned for specific applications. Finally, aesthetic research has primarily focused on colour matching, with additional analysis based on personality, photographic content, and direct visual feature computation. The subjective nature of aesthetics and the lack of clear principles or standards make it a challenging research area that lacks a strong framework or benchmarking system.
  
\section{Generation Tasks}
\label{sec:generation tasks}
Graphic design needs necessitate elements with separate transparent backgrounds. Thus, we focus primarily on vector shape generation and the vectorization of artistic imagery. For ease of discussion, we categorize text element generation into the generation of text itself and the rendering of text within a scene. Meanwhile, we introduce works in layout generation and layout-based image generation. Finally, we address research focused on aesthetic refinement.

\vspace{-3mm}
\subsection{Non-text Element Generation}
\vspace{-1mm}
\label{sec:non text ge}
\subsubsection{SVG Generation.}
\label{sec:non text ge ve}
SVG can be encoded as the sequence of 2d points connected by the parametric curves, making the seq2seq model straightforward for encoder/decoder basis~\cite{deepsvg2020,ha2017sketchrnn,lopes2019learned,wang2021deepvecfont,iconshop2023}. SketchRNN~\cite{ha2017sketchrnn} was a pioneer in employing LSTM-based VAEs for learning to draw strokes, representing sketches as sequences of pen positions and states. SVG-VAE~\cite{svgvae2019} involves a two-stage training process that begins with an image-based VAE, followed by training a decoder to predict vector parameters from the latent variables. BézierSketch~\cite{das2020beziersketch} focuses on generating Bézier curves, offering enhanced control over the graphical forms of sketches. DeepSVG~\cite{deepsvg2020}, a hierarchical autoencoder designed to learn representations of vector paths, contributes to the structural complexity of vector graphics. These methods heavily rely on datasets in vector form, which limits their generalization capabilities and their capacity to synthesize complex vector graphics. IconShop trains a BERT model for text-conditioned SVG generation of icons, but their method is restricted to using paths~\cite{iconshop2023}. 

Instead of directly learning an SVG, another method of vector synthesis optimizes towards a paired raster image during training. ~\cite{li2020differentiable} observed vector graphics rasterization is differentiable after pixel prefiltering. Conditioned on this finding, the authors introduce a differentiable rasterizer that offers two prefiltering options: an analytical prefiltering technique and a multisampling anti-aliasing technique. The analytical variant is faster but can suffer from artefacts such as conflation. The multisampling variant is still efficient and can render high-quality images while computing unbiased gradients for each pixel with respect to curve parameters. Thanks to this work enables the supervision of the SVG generation under the guidance of a raster image. In other words, different from image generation methods that traditionally operate over vector graphics require a vector-based dataset, recent work has demonstrated the use of differentiable renderers to overcome this limitation~\cite{ma2022towards,reddy2021im2vec,shen2021clipgen,song2023clipvg,su2023marvel,xing2023diffsketcher}. CanvasVAE defines vector graphic documents through a multi-modal set of attributes, using variational autoencoders to integrate diverse graphical components~\cite{yamaguchi2021canvasvae}. Im2Vec developed a method that employs a differentiable rasterization pipeline to generate complex vector graphics from raster training images~\cite{reddy2021im2vec}. Furthermore, recent advances in visual text embedding and contrastive language-image pre-training models have enabled a number of successful methods for synthesizing sketches, including CLIPDraw and CLIPasso~\cite{frans2022clipdraw, vinker2022clipasso}. In addition to using CLIP distance, VectorFusion~\cite{jain2023vectorfusion} and DiffSketcher~\cite{xing2023diffsketcher} combine a differentiable renderer with a text-to-image diffusion model for vector graphics generation. This type of method utilizes Score Distillation Sampling loss based on a Text-to-Image (T2I) diffusion model for optimizing SVG to align with text prompts across various applications such as fonts, vector graphics, and sketches~\cite{iluz2023word,gal2024breathing,xing2024svgdreamer}. However, due to the lack of geometric constraints, they often lead to path intersections or jagged effects. By adding geometric constraints to a Text-to-Vector (T2V) generation pipeline that optimizes local neural path representation, high-quality SVG graphics generation is achieved~\cite{zhang2024text}.

\subsubsection{Vectorization of Artist-Generated Imagery.}
\label{sec:non text ge i2v}
Image vectorization is another alternative way to directly obtaining the bitmap from imagery. Traditional vectorization techniques primarily depend on segmentation or edge detection to group pixels into larger regions, subsequently fitting vector curves and region primitives to these segments~\cite{sun2007image,xia2009patch}. Challenges include aligning patch boundaries and automating mesh generation~\cite{lai2009automatic,xia2009patch}. In contour-based vectorization, simpler elements such as lines, circles, and Bézier curves represent discontinuity sets in piecewise constant images, often including silhouettes and pixel art~\cite{zhang2009vectorizing,sykora2005sketching}.
To better fit piecewise smooth vector curves to raster boundaries, 
~\cite{bessmeltsev2019vectorization} propose an image vectorization method based on mathematical algorithms for frame field processing.
PolyFit\cite{dominici2020polyfit} approximates piecewise smooth vector curves to raster boundaries with coarse polygons, considering perceptual cues and simplicity. LIVE~\cite{ma2022towards} and SAMVG~\cite{zhu2024samvg} employ a layer-wise optimization framework that significantly improves vectorization quality. Chen et al.\cite{chen2023editable} explore the assembly of simple parametric primitives within a neural network for geometric abstraction. SuperSVG~\cite{hu2024supersvg} focuses on decomposing the input into superpixels for optimized reconstruction and detail refinement.

\vspace{-3mm}
\subsection{Text Element Generation}
\vspace{-1mm}
\label{sec:text ge}
\subsubsection{Artistic Typography Generation.}
\label{sec:text ge ar}
One of the main directions in text element generation is font style learning. Traditional methodologies centred on explicit shape modelling and statistical learning techniques to craft font glyphs of calligraphy, predominantly targeting elements such as strokes and radicals\cite{zhou2011easy, lian2016automatic}. Research efforts focused on emulating traditional calligraphic styles using hierarchical models and texture transfer techniques~\cite{phan2015flexyfont, tenenbaum1996separating, goda2010texture, murata2016japanese}. Deep learning has markedly increased the flexibility and realism in font creation; researchers have utilized image translation methods for font generation and explored font style learning in one-shot and few-shot settings. ~\cite{tian2017zi2zi} was the first to adopt GANs to automatically generate a Chinese font library by learning a mapping from one style font to another, and DC-Font~\cite{jiang2017dcfont} also addresses the font feature reconstruction and handwriting synthesis problems through adversarial training. However, these methods operate under supervised learning and necessitate a large volume of paired data. Some methods employ auxiliary annotations (e.g., stroke and radical decomposition) to enhance generation quality further.  RDGAN~\cite{huang2020rd} proposes a radical extraction module to extract rough radicals. To facilitate the automatic synthesis of new fonts more easily, some works follow unsupervised methods to separately obtain content and style features and then fuse them in a generator to produce new characters~\cite{zhang2018separating, gao2019artistic, xie2021dg}. Concurrently, other works leverage auxiliary annotations to make the models cognizant of the specific structure and details of glyphs~\cite{kong2022look, tang2022few}. Most recently, Diff-Font~\cite{he2024diff} represents a pioneering effort to use a diffusion model, treating it as a conditional generation task to manage content through predefined embedding tokens while extracting the desired style from a one-shot reference image.

Another significant line is transferring artistic styles of color and texture onto new glyphs. Yang et al.~\cite{yang2017awesome} pioneered text effect transfer by enabling the migration of effects from a stylized text image to a plain one. Following this, ~\cite{men2018common} developed a general framework for user-guided texture transfer, applicable to a variety of tasks, including transforming doodles into artworks, editing decorative patterns, generating texts with special effects, controlling effect distribution in-text images, and swapping textures. The following research in this domain revolves around exploring and enhancing techniques for separating, transforming, and recombining text styles and content. Research has progressively evolved to address various aspects of style and content encoding, mixing, and decoding. \cite{azadi2018multi} pioneered the application of deep networks to text effect transfer, focusing on combining font and text effects.  \cite{chen2023style} addressed specific issues like stroke adhesion and text clarity, while \cite{xue2023art} tackled data scarcity through synthetic data generation. The field has recently seen innovative approaches leveraging diffusion models for diverse style support and interactive generation, culminating in Wang et al. \cite{wang2023anything}'s method for generating artistic fonts using a text-to-image diffusion model.

In semantic typography, the primary goal is to enable the integration of artistic expression with legibility while embedding semantic meanings into typographic designs. Xu et al. \cite{xu2007calligraphic} pioneered an interactive method for creating calligrams that warp letters to fit within specific regions of an image, aligning them semantically with the visual content, albeit sometimes at the cost of readability. Building on the need for clarity, Zou et al. \cite{zou2016legible} refined guidelines for glyph deformation through a crowd-sourced study, aiming to improve the readability of automated letter layouts. To further personalize and enrich typography, Zhang et al. \cite{zhang2017synthesizing} introduced a framework that allows users to influence glyph structure interactively, incorporating a semantic-shape similarity metric and optional structural optimization techniques to enhance both aesthetics and integrity. Advancements continued with Iluz et al. \cite{iluz2023word} who modified letter geometry based on semantic meanings and employed advanced rendering techniques to ensure high-quality visualization across sizes. Finally, Tanveer et al. \cite{tanveer2023ds} leveraged large language models and unsupervised generative models to synthesize stylized fonts with embedded semantic meanings.

\subsubsection{Visual Text Rendering.}
\label{sec:text ge visual}
Text rendering target to render the text characters in imagery. Methodologies in this area have been extensively researched to address the issue of visual inconsistency often observable when text is merely superimposed onto images. Notable approaches like SynthText~\cite{gupta2016synthetic}, VISD~\cite{zhan2018verisimilar}, SynthText3D~\cite{liao2020synthtext3d}. Despite these technological advancements, the field continues to face significant challenges in achieving accurate text rendering and ensuring visual coherence with the surrounding environment, primarily due to the limited diversity in background datasets utilized for training and synthesis. Most existing research efforts~\cite{chen2024textdiffuser} have concentrated on the precise visual rendering of text in English. However, initiatives like AnyText~\cite{tuo2023anytext} show only moderate success in rendering texts in other languages such as Chinese, Japanese, and Korean. This is largely attributed to the challenges in gathering high-quality data and the constraints of training models on a limited dataset comprising merely 10,000 images across five languages. Given the extensive array of characters in these non-English languages, such a dataset size proves inadequate for comprehensively addressing the task of multilingual visual text rendering. Furthermore, contemporary commercial image generation models like DALL·E3, Imagen3, Stable Diffusion 3~\cite{esser2024scaling}, and Ideogram 1.01 have demonstrated underwhelming performance in multilingual text rendering tasks.

Recent research has focused on enhancing text rendering accuracy by integrating large-scale language models such as T5, used by platforms like Imagen \cite{saharia2022photorealistic}. Studies suggest that character-aware models like ByT5 \cite{xue2022byt5} offer substantial advantages over character-blind models such as T5 and CLIP \cite{radford2021learning} in terms of text rendering accuracy. Innovations such as GlyphDraw \cite{ma2023glyphdraw} introduce frameworks for precise control over character generation, incorporating features like auxiliary text locations and glyph characteristics. TextDiffuser \cite{chen2024textdiffuser} uses a Layout Transformer to enhance knowledge of text arrangement and integrates character-level segmentation masks for higher accuracy. GlyphControl \cite{yang2024glyphcontrol} and Diff-Text \cite{zhang2024brush} refine the approach by facilitating explicit learning of text glyph features and using rendered sketch images as priors for multilingual generation, respectively. Meanwhile, GlyphOnly \cite{li2024first}, which uses glyphs and backgrounds for accurate rendering and consistency control, is equipped with an adaptive strategy for exploring text blocks in small-scale text rendering.

\vspace{-3mm}
\subsection{Layout Generation}
\vspace{-1mm}
\label{sec:layout ge}
\subsubsection{Automatic Layout Generation}
\label{sec:layout de}
Layout can be created by selecting a template that best fits the content~\cite{lee2010designing,dayama2020grids,o2015designscape}. However, such a predefined, constrained set of templates can rarely accommodate the vast diversity of graphic design layouts. Many works studied on creating a layout according to the given elements for graphic design such as UI design\cite{todi2016sketchplore}, advertisement design~\cite{lee2020neural, li2020attribute}, website~\cite{jing2023layout}, book covers~\cite{zhang2021towards}, magazine design~\cite{yang2016automatic}, and poster design~\cite{chai2023two,guo2021vinci}, among others. Early research on design layouts primarily utilized templates, exemplars, and heuristic design rules~\cite{damera2011probabilistic, hurst2009review, o2014learning, tabata2019automatic}. These methods, which often required professional design knowledge, leveraged predefined templates or heuristic-based rules but struggled to address the diversity and complexity of design requirements effectively. Subsequent developments introduced techniques such as saliency maps~\cite{bylinskii2017learning} and attention mechanisms~\cite{pang2016directing}. These methods were designed to assess the visual importance within graphic designs, track user attention, and enhance understanding of how users engage with visual elements, marking a significant step towards understanding the dynamics of visual interaction in layouts. Neural networks enable researchers to derive design principles from extensive datasets. CanvasVAE~\cite{yamaguchi2021canvasvae} introduced a VAE-based architecture for unconditionally generating vector graphic documents. Following this, LayoutGAN\cite{li2019layoutgan} further refined this approach by incorporating user-specified constraints into layout generation. LayoutDM\cite{chai2023layoutdm} employs DDPM to handle geometric parameters in continuous spaces while introducing category information as a condition. LayoutDiffusion~\cite{zhang2023layoutdiffusion} treats both geometric parameters and category information as discrete data. LDGM~\cite{hui2023unifying} proposes to decouple the diffusion processes to improve the diversity of training samples and learn the reverse process jointly. These methods are learned from the vector domain. A recent work~\cite{shabani2024visual} combines the advantages from both bit vector and raster image spaces by proposing a dual diffusion model for design layout generation.

In addition, layout generation in raster images has evolved into two directions: content-agnostic and content-aware layout generation. Content-agnostic layout generation focuses on generating layouts without a predefined content structure. Techniques such as LayoutVAE~\cite{jyothi2019layoutvae}, which utilizes a VAE, and others employing auto-regressive models~\cite{gupta2021layouttransformer, arroyo2021variational, kong2022blt} or diffusion models~\cite{inoue2023layoutdm, zhang2023layoutdit} have been prominent. DLT~\cite{levi2023dlt} further advances this by integrating discrete and continuous data in a diffusion layout transformer. Content-aware layout generation integrates specific visual and textual content into the layouts, aiming to create more contextually relevant designs. Early innovations include Content-GAN~\cite{zheng2019content}, which was the first to combine visual and textual elements. Subsequent models like and ICVT~\cite{cao2022geometry} employ transformer-based networks and conditional VAEs, respectively, to enhance content integration. PosterLayout~\cite{hsu2023posterlayout} uses a CNN-LSTM network focusing on saliency maps. LayoutDETR~\cite{yu2025layoutdetr} leverages a Detection Transformer approach, integrating GAN and VAE technologies and utilizing pre-trained visual and textual encoders for feature extraction. 

Furthermore, layouts, which can be encoded in formats such as XML or JSON, are ideally suited for processing by pre-trained LLMs. To this end, a series of works utilize the paradigm of code generation + LLM~\cite{tang2023layoutnuwa}. LayoutGPT~\cite{feng2024layoutgpt} utilizes in-context visual demonstrations in CSS structures to enhance the visual planning capabilities of GPT-3.5/4 for generating layouts from textual conditions. MuLan~\cite{li2024mulan} iteratively plans the layout of an image by deconstructing the text prompt into a sequence of sub-tasks with an LLM, then revises the image at each step based on feedback from a vision-language model. TextLap~\cite{chen2024textlap} enables users to generate layout designs based on natural-language descriptions.  LayoutPrompter~\cite{lin2024layoutprompter} introduces a training-free approach by leveraging Retrieval-Augmented Generation to enhance the in-context learning capabilities of GPT~\cite{brown2020language}, dynamically sourcing examples from a dataset. However, this retrieval-centric strategy is limited to open-domain generation. These works often overlook the visual domain features or convert them into hard tokens before inputting them into LLMs, which can result in significant information loss.

\subsubsection{Glyph Layout Generation}
\label{sec:layout guided}
\cite{wang2022aesthetic} were the first to propose this task. The synthesized layouts of glyphs must consider fine-grained details, such as avoiding the collision of strokes from different glyphs. Furthermore, the placement trajectories of characters should follow a correct reading order (e.g., left to right and top to bottom for English) and possess diverse styles simultaneously, challenges that non-sequence generation models struggle to handle. To address these issues, the authors introduced a dual-discriminator module designed to capture both the character placement trajectory and the rendered shape of the synthesized text logo. However, it faces challenges in designing layouts for long text sequences, adapting to user-defined constraints, and providing diverse layout designs due to the limited amount of training data. In response, GLDesigner \cite{he2024gldesigner} proposed a vision-language model-based framework that generates content-aware text logo layouts by integrating multimodal inputs with user constraints. This study also includes the creation of two extensive text logo datasets, which are five times larger than any existing public datasets. In addition to geometric annotations, such as text masks and character recognition, comprehensive layout descriptions in natural language format have been provided to enhance reasoning capabilities. Although this model indeed improves the fidelity of generated visual text, it generally falls short in rendering longer textual elements. Lakhanpal et al. \cite{lakhanpal2024refining} introduce a training-free framework to enhance two-stage generation approaches focusing on generating images with long and rare text sequences.

\vspace{-3mm}
\subsection{Colorization}
\vspace{-1mm}
\label{sec:color}
Image Colorization is the process of converting grayscale images, including manga~\cite{ouyang2021interactive}, line art~\cite{yuan2021line}, sketches~\cite{zhang2021line,wang2023region}, and grayscale photographs~\cite{zabari2023diffusing}, into their full-color versions. Various techniques are employed to guide the colorization process, such as scribbles~\cite{
dou2021dual,yun2023icolorit,zhang2021user}, reference images~\cite{bai2022semantic,li2021globally,li2022style,li2022eliminating,wang2023unsupervised,wu2023self,wu2023flexicon,zhang2022scsnet,zou2024lightweight}, color palettes~\cite{wang2022palgan,wu2023flexicon}, and textual descriptions~\cite{chang2023coins,weng2023cad,zabari2023diffusing,zhang2023adding}. Scribbles are used to provide intuitive and spontaneous color hints through freehand strokes. The Two-stage Sketch Colorization~\cite{zhang2018two} incorporates a CNN-based system that first applies preliminary color strokes to the canvas, which are later refined to improve color accuracy and detail. Colorization using reference images involves transferring color schemes from an image with similar elements, scenes, or textures. Methods based on stroke application, or edit propagation, allow users to manually introduce color alterations using strokes that are algorithmically extended across the image based on criteria like color similarity and spatial relationships. These methods are invaluable for targeted color adjustments and preserving the authentic appearance of the image. Developments in this field have introduced neural network-driven techniques that automate edit propagation across comparable image structures~\cite{cheng2015deep, endo2016deepprop}. Palette-based techniques aim to distill the essential color scheme of an image into a select group of representative colors, thereby reducing and abstracting the rich diversity of colors present. Innovations by Chang et al. involved adapting a K-means clustering algorithm to extract a color palette, which laid the groundwork for later advancements. Palette-based models~\cite{chang2015palette} utilize the selected palette as a stylistic guide to influence the overall color theme of the image. Example-based approaches, or style transfer, utilize existing images as templates to guide the recoloring effort, allowing for the transfer of stylistic color elements from one image to another, a process enhanced through the use of CNNs and GANs~\cite{zhang2016colorful,vitoria2020chromagan}. With the rise of diffusion models, textual descriptions have become a pivotal tool for image generation, and thus play a significant role in image colorization. Text-based guidance employs descriptions of desired color themes, object colors, or mood. ControlNet~\cite{zhang2023adding} integrates additional trainable modules into pre-existing text-to-image diffusion models~\cite{rombach2022high}, tapping into the inherent capabilities of diffusion models for colorization tasks.

\vspace{-3mm}
\subsection{Summary}
\vspace{-1mm}
\label{sum2}
SVGs use geometric primitives like Bezier curves, polygons, and lines, making them well-suited for representing visual concepts in a structured, scalable format. DiffVG~\cite{li2020differentiable} allows seamless transitions between raster and vector images. Research in visual text generation can be divided into three main categories: basic text generation, artistic text generation, and text rendering in natural scene images. (1) Basic text generation primarily deals with font transfer, especially for complex scripts such as Chinese or Korean. The focus extends beyond simple generation to include text segmentation into its constituent parts like radicals and strokes, which helps guide the learning process more effectively. (2) Artistic text generation encompasses two main tasks: artistic style transfer and semantic text generation. Both areas use the same basic generation models but tackle different challenges. Artistic style transfer focuses on separating the style and content of text. Semantic text generation, conversely, involves self-deformation of text to maintain readability and aesthetic appeal during automatic re-layout. (3) The rendering of text in natural scene images is particularly challenging due to the need for high clarity and accuracy in diverse visual contexts. To address these challenges, researchers utilize large pre-trained models like Google's T5 series, which is adept at character perception.

The study of layout is a prominent topic. Traditional template-based methods, prevalent in early research, often struggle to encapsulate design rules effectively. Contemporary training-based approaches are categorized based on the type of data used: bitmap data and raster image data. Additionally, a recent innovative work~\cite{shabani2024visual} integrates a dual diffusion model encompassing bitmap and raster images.
Furthermore, another significant advancement is the encoding of layouts in formats such as XML or JSON, which are highly compatible with processing by pre-trained LLMs. Recent studies have begun to conceptualize layout generation as a language reasoning or planning task. Further developments include the integration of visual information.

Aesthetic research specifically focusing on recolorization. Traditionally, coloring methods predominantly utilized rule-based systems complemented by color propagation strategies. CNNs were trained on paired grayscale and color images, facilitating the learning of direct mappings from grayscale inputs to their colorized counterparts. Further developments saw the introduction of GANs. In this setup, a generator network learns to create color images that emulate the real colors found in the dataset, while a discriminator network ensures these colorizations are consistent with the original grayscale images. Despite the success of CNNs, their limitations in capturing long-range dependencies prompted researchers to explore other architectures. The introduction of transformers, known for their ability to handle global contexts, brought new opportunities. The most recent advancements involve diffusion models, which integrate pre-trained models to better understand the semantics of color. These models offer guidance during the colorization process by leveraging learned representations of color and its contextual significance, leading to more nuanced and semantically coherent outputs.

\section{Present and Future}
\label{cha}
In this section, we examines the current trend of AIGD, identifies key challenges, and outlines future directions, prevailing research trend of addressing comprehensive problems through holistic solutions.

\vspace{-3mm}
\subsection{MLLM for Graphic Design}
\vspace{-1mm}

\begin{figure*}[t]
  \begin{center}
  \includegraphics[width=1\linewidth]{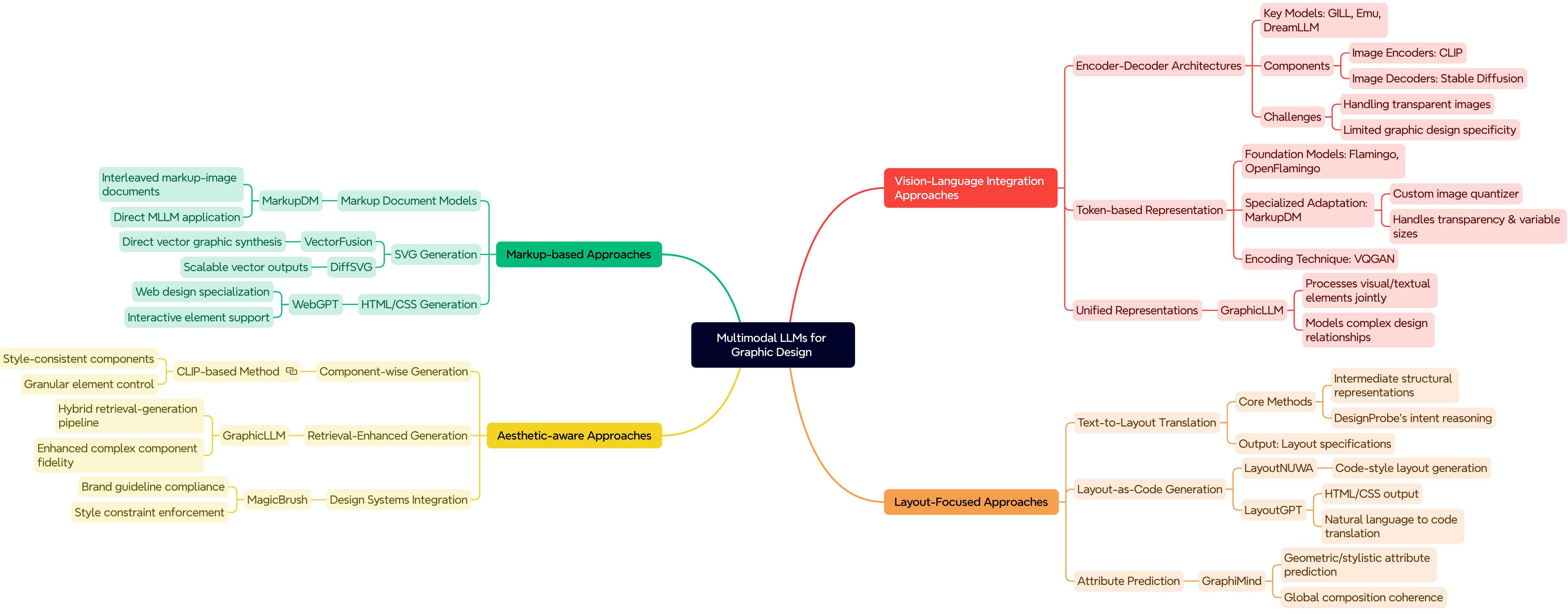}
  \end{center}
  \vspace{-4.0mm}
  \caption{Overview of existing works in Multimodal LLM for graphic design.} 
  \label{fig11}
  \vspace{-5mm}
\end{figure*}

Recent advancements in multimodal LLMs have shown promising applications in graphic design tasks. As shown in Fig.~\ref{fig11}, these approaches can be categorized into several key directions based on their architectural design and application focus.

\textbf{Vision-Language Integration Approaches} which connect LLMs with pre-trained image components:

\begin{itemize}
    \item \textbf{Encoder-Decoder Architectures:} Models such as DreamLLM \cite{dong2023dreamllm} integrate LLMs with pre-trained image encoders (e.g., CLIP \cite{radford2021clip}) and decoders (e.g., Stable Diffusion \cite{rombach2022stablediffusion}). While powerful for general image generation, these approaches face challenges with transparent images common in graphic design \cite{kikuchi2024multimodal}. OpenCLOE~\cite{inoue2024opencole} begins by translating user intentions into a design plan using GPT-3.5 and in-context learning. Then, the image and typography generation modules synthesize design elements according to the specified plan and the graphic renderer assembles the final image.
    
    \item \textbf{Token-based Representation:} An alternative approach represents images as discrete tokens \cite{alayrac2022flamingo, awadalla2023openflamingo, liu2023visual}. This method encodes images into token sequences via image quantizers like VQGAN \cite{esser2021vqgan}. The MarkupDM approach \cite{kikuchi2024multimodal} adapts this methodology specifically for graphic design by developing a custom quantizer that handles transparency and varying image sizes.
    
    \item \textbf{Unified Models:} GraphicLLM \cite{cheng2024graphicllm} proposes a multimodal model that processes both visual and textual design elements within a unified framework, addressing the complex relationships.
\end{itemize}

\textbf{Layout-Focused Approaches} which leverage LLMs specifically for layout generation tasks:

\begin{itemize}
    \item \textbf{Text-to-Layout Translation:} \cite{qu2023layoutllm} utilize LLMs to translate descriptions into intermediate structural representations that guide subsequent layout generation. DesignProbe \cite{lin2024designprobe} extends this by introducing a reasoning mechanism where LLMs analyze design intent before generating structured layout specifications.
    
    \item \textbf{Layout-as-Code Generation:} LayoutNUWA \cite{wang2024layoutnuwa} treats layout generation as a code generation task, leveraging the programming capabilities of LLMs. Similarly, LayoutGPT \cite{feng2023layoutgpt} functions as a layout generator by producing HTML/CSS code from textual prompts.
    
    \item \textbf{Attribute Prediction:} GraphiMind \cite{huang2024graphimind} employs MLLMs to predict geometric and stylistic attributes for design elements while maintaining global coherence across the entire composition.
\end{itemize}

\textbf{Aesthetic-aware Approaches} aim to address multiple aspects of design simultaneously:

\begin{itemize}  
    \item \textbf{Component-wise Generation:} \cite{gao2024clip} propose a method that leverages CLIP embeddings to generate design components that maintain stylistic consistency. VASCAR~\cite{zhang2024vascar} is large vision-language models-based content-aware layout generation. Design-o-meter~\cite{goyal2024design} is the first work to score and refine designs within a unified framework by adjusting the layout of design elements to achieve high aesthetic scores.
    \item \textbf{Retrieval-Enhanced Generation:} GraghicLLM \cite{cheng2024graphicllm} combines generative capabilities with retrieval mechanisms to leverage existing design elements, achieving higher fidelity results for complex graphic components.
    
    \item \textbf{Design Systems Integration:} MagicBrush \cite{zhang2024magicbrush} integrates with design systems to ensure generated elements conform to established brand guidelines and stylistic constraints.
\end{itemize}

\textbf{Markup-based Approaches.} involve representing designs as markup language:

\begin{itemize}
    \item \textbf{Markup Document Models:} MarkupDM \cite{kikuchi2024multimodal} introduces a novel approach treating graphic designs as interleaved multimodal documents consisting of markup language and images. This representation allows direct application of multimodal LLMs to graphic design tasks.
    
    \item \textbf{SVG Generation:} VectorFusion \cite{jain2023vectorfusion} focus on generating vector graphics (SVG) directly, addressing the scalability advantages needed for professional graphic design workflows.
    
    \item \textbf{HTML/CSS Generation:} WebGPT \cite{lee2023webgpt} generates web-based designs by producing HTML and CSS code, demonstrating the potential of code-centric approaches for interactive designs.
\end{itemize}

\vspace{-3mm}
\subsection{Existing Challenges}
\vspace{-1mm}
\begin{figure*}[t]
  \begin{center}
  \includegraphics[width=1\linewidth]{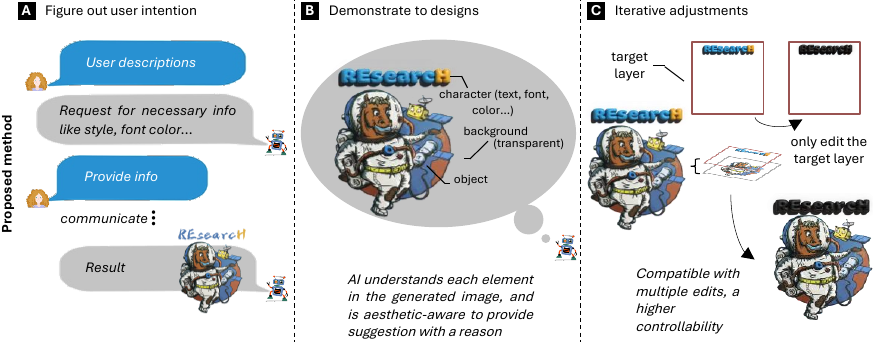}
  \end{center}
  \vspace{-4mm}
  \caption{Demonstration AIGD, including autonomous in A to figure out user intention, interpretable in B the explanation can be provided to demonstrate designs and multi-layered in C when edit generated designs.} 
  \label{fig12}
  \vspace{-5.5mm}
\end{figure*}

Despite remarkable advances in AIGD, significant technical barriers that limit practical applications persist. As illustrated in Fig.~\ref{fig12}, conventional approaches suffer from three fundamental limitations: inadequate user intention understanding, limited interpretability, and insufficient layer control. These challenges reflect deeper systemic issues within AI architectures requiring targeted research attention.

The problem of inadequate user intention understanding (Fig.\ref{fig12} A) represents a fundamental cognitive-computational gap in current systems. While text-to-image models have advanced significantly, they fundamentally operate through statistical pattern matching rather than the semantic understanding of design requirements. This semantic-representational decoupling manifests when converting textual briefs to visual styles, where the design intention encoding $\mathbf{h}I = \text{LLM}\theta(\mathbf{x}{\text{text}}) \oplus \text{ViT}\phi(\mathbf{x}_{\text{image}})$ yields demonstrably lower style consistency than human designers in controlled trials. The challenge extends beyond simple prompt engineering to the more complex problem of modelling designers' cognitive processes when translating abstract requirements into concrete visual decisions. Current multimodal approaches attempt to bridge this gap through joint embeddings but fail to capture the nuanced contextual reasoning that experienced designers apply to interpret client needs, audience expectations, and brand guidelines simultaneously \cite{kikuchi2024multimodal, cheng2024graphicllm}. This intention-representation gap becomes particularly evident in iterative feedback scenarios, where AI systems struggle to incorporate targeted revisions without regenerating entire compositions. This limitation significantly reduces their utility.

The interpretability challenge (Fig.\ref{fig12} B) reflects a deeper epistemological problem in AI-driven design: the inability to articulate design rationale in terms that align with established design principles and practices. Current systems provide generic explanations that lack specificity regarding composition decisions, color harmony, typographic choices, and other critical design elements. This limitation stems partly from how design knowledge is encoded during training—predominantly through implicit pattern recognition rather than explicit design theory. The catastrophic performance degradation when transitioning from atomic to composite tasks represents a direct consequence of this interpretability deficit. Models cannot effectively decompose complex tasks into meaningful subproblems or explain the interrelationships between design elements without an explicit representational framework for design principles. Recent work in design rationale extraction has attempted to explain generated designs retroactively. Still, these post-hoc rationalizations often fail to reflect the actual generative process or provide actionable insights for refinement. Typography presents an additional interpretability challenge—while font generation models have achieved impressive stylistic accuracy, they frequently fail to balance aesthetic considerations with functional requirements like readability across contexts, proper kerning, and linguistic nuance.

The multiple layers and iterative editing problem (Fig.\ref{fig12} C) reveals fundamental architectural limitations in current generative models when applied to professional graphic design workflows. Unlike the photography-oriented generation, graphic design requires precise control over individual elements, relationships, and layer hierarchies. Conventional methods struggle with layer-specific editing—modifications to one element often unintentionally affect others, as evident in tools like TurboEdit and Flux.1. This limitation stems from pre-trained image encoders and decoders that inadequately support transparent images and multi-layered compositions \cite{kikuchi2024multimodal}. The technical challenge extends beyond simple transparency support to the more complex problem of maintaining semantic consistency across layers while enabling targeted modifications. Standard diffusion models and transformers operate on flattened representations that fail to preserve the logical independence of design elements.  This represents a fundamental tension between maintaining high-fidelity visual details and producing clean, scalable vector representations—a tradeoff that simultaneously impacts editability, rendering performance, and file size.

Beyond these three primary challenges, a fourth systemic limitation has emerged in recent research: contextual consistency maintenance across design artifacts. Professional graphic design rarely involves isolated images, instead requiring coherent visual systems spanning multiple formats and applications while maintaining brand identity. Current AI approaches treat each generation as an independent task, lacking the architectural components to model and maintain cross-artifact consistency. This limitation becomes particularly problematic in comprehensive design systems where visual elements must adapt to different contexts (responsive web design, print media, environmental applications) while preserving core identity elements. The computational challenge involves maintaining a persistent design representation that can flexibly adapt to new constraints without sacrificing fundamental stylistic principles—a problem that remains largely unaddressed in current research. These challenges collectively point to a need for more sophisticated architectural approaches that better align with the cognitive processes, theoretical foundations, and practical workflows of professional graphic design. While recent multimodal models show promising capabilities in specific domains, bridging the gap to comprehensive design assistance requires addressing these fundamental limitations through targeted research in representation learning, explainable generative processes, compositional reasoning, and layered editing paradigms.

\vspace{-3mm}
\subsection{Potential Directions}
\vspace{-1mm}
Recent trends in AI for graphic design suggest several promising research directions that warrant further investigation. This section explores potential avenues for advancing the field, focusing on developing unified approaches and addressing specific challenges in design understanding and generation.

\textbf{Towards Unified End-to-End Models.} Recent advances in MLLMs demonstrate the feasibility of employing unified end-to-end solutions~\cite{le2024one} for AIGD tasks. These models would integrate multimodal intent understanding, high-quality visual element generation, and knowledge-enhanced layout reasoning within a single framework. Such integration aligns with current academic trajectories in multimodal learning and offers a promising pathway for comprehensively addressing the complex challenges of graphic design automation.

\begin{itemize}
\item \textbf{Multimodal Intent Understanding.} Current multimodal models integrating dialogue and visual recognition provide a foundation for intent understanding but require significant enhancement in several key areas: 1) Graphic design presents unique challenges with artistic images featuring diverse fonts and complex layouts that exceed the capabilities of general-purpose recognition systems. 2) 3D designs, text with special effects (overlapping, bending, distortion), and artistic typography demand specialized recognition approaches. 3) Enhanced communicative abilities in large language models are needed to translate ambiguous user inputs into coherent, actionable design specifications.

\item \textbf{Knowledge-Enhanced Layout Reasoning.} The computational representation of abstract design principles presents significant challenges. Drawing inspiration from advanced reasoning models like OpenAI o1, research should focus on: 1) Encoding established design theories within computational frameworks. 2) Developing inference mechanisms that can apply these principles contextually. 3) Creating evaluation metrics that align with human aesthetic judgment. 4) Building models that can explain their layout decisions with reference to design principles.

\item \textbf{High-Quality Visual Element Generation.} Layer diffusion techniques show promise for creating images with transparent backgrounds—a critical requirement for graphic design. However, text generation capabilities require substantial improvement, particularly for artistic typography, where models like Flux.1 demonstrate potential but insufficient fidelity. Meanwhile, LLM-guided approaches for generating vector graphics, exemplified by tools like SVGDreamer, offer precision and scalability advantages. Research should focus on enhancing text rendering and incorporating deeper reasoning about design principles. Finally, models capable of a seamless transition between raster and vector formats could revolutionize workflow efficiency by offering the advantages of both paradigms, as suggested by~\cite{shabani2024visual}.
\end{itemize}

\textbf{Research in Sub-directions.} Beyond unified models, several specialized research directions show particular promise: 1) Developing encoders specifically trained on graphic design elements could substantially improve the representation of design-specific features. Unlike general-purpose visual encoders, design-specific approaches would prioritize typographical feature representation, layout structure encoding, color harmony, and palette relationships. 2) Interactive and Collaborative Design Systems enable iterative refinement and feedback loops between the designer and AI, focusing on turn-taking mechanisms for collaborative design, interpretable design suggestions, learning from designer feedback and preserving creative agency while enhancing productivity. 3) Design Rationale Understanding Models that capture the underlying reasoning in design decisions, rather than just visual patterns, represent a critical frontier~\cite{lin2024designprobe}, which involves inferring design intentions from examples, reverse-engineering design decisions, representing the relationship between design goals and visual implementations and learning from design critique and evaluation. 4) 
Improved mechanisms for transferring design styles between different modalities offer significant potential for design consistency and efficiency~\cite{zhou2024transfusion}. 5) The vector graphics domain, particularly SVG, remains underexplored despite its importance in graphic design. Recent work by~\cite{wang2024visually} introduces Primal Visual Description (PVD), a textual representation that translates SVGs into abstractions comprising primitive attributes (shape, position, measurement) and their values. Research in this domain should explore the integration of these structured representations with generative and reasoning capabilities, potentially offering precision advantages over raster-based approaches for graphic designs.

The intersection of AI and graphic design presents rich research opportunities that bridge visual generation, multimodal understanding, and design reasoning. The emerging field of multimodal LLMs for graphic design demonstrates significant promise for automating and enhancing design workflows, particularly as these models continue to improve in handling the unique characteristics of design documents, including transparency, layout constraints, and stylistic coherence. Progress will require interdisciplinary approaches to transform design workflows while preserving and enhancing human creative agency.

\section{Conclusion}
\label{con}
This survey has comprehensively reviewed AI's state-of-the-art methods and applications in Graphic Design, categorizing them into perception and generation tasks. We have explored various subtasks, including non-text element perception, text element perception, layout analysis, aesthetic understanding, and the generation of non-text elements, text elements, layouts, and colors. Integrating large language models and multimodal approaches has become a pivotal trend, enabling more holistic and context-aware design solutions. However, several challenges persist, such as the need to better understand human intent, improved interpretability of AI-generated designs, and enhanced control over multi-layered compositions. Future research should develop unified end-to-end models integrating multimodal understanding, high-fidelity generation, and design reasoning.

\bibliographystyle{unsrt} 
\bibliography{1}  

\end{document}